\documentclass[10pt,twocolumn,letterpaper]{article}

\usepackage{iccv}      
\usepackage{multirow}
\usepackage{colortbl}
\usepackage{graphicx}
\usepackage{xcolor}
\usepackage{pifont}

\definecolor{iccvblue}{rgb}{0.21,0.49,0.74}
\usepackage[pagebackref,breaklinks,colorlinks,allcolors=iccvblue]{hyperref}
\usepackage{bm}
\def\paperID{15572} 
\def\confName{ICCV}
\def\confYear{2025}

\title{DistillDrive: End-to-End  Multi-Mode Autonomous Driving Distillation by Isomorphic Hetero-Source Planning Model}

\author{Rui Yu$^{1}$\qquad Xianghang Zhang$^{2}$ \qquad Runkai Zhao$^{3}$\qquad Huaicheng Yan$^{1}$\textsuperscript{\dag}\qquad Meng Wang$^{1}$\\
$^{1}$ East China University of Science and
Technology\quad$^{2}$ SenseAuto Research\quad$^{3}$ The University of Sydney\\
\tt\small y80220166@mail.ecust.edu.cn\quad zhangxianghang@senseauto.com\quad  rzha9419@uni.sydney.edu.au\quad \\
\tt\small hcyan@ecust.edu.cn\quad mengwang@ecust.edu.cn\quad$^\dag$ Corresponding author
}

\begin{document}
\maketitle
\begin{abstract}
End-to-end autonomous driving has been recently seen rapid development, exerting a profound influence on both industry and academia.
However, the existing work places excessive focus on ego-vehicle status as their sole learning objectives and lacks of planning-oriented understanding, which limits the robustness of the overall decision-making prcocess.
In this work, we introduce DistillDrive, an end-to-end knowledge distillation-based autonomous driving model that leverages diversified instance imitation to enhance multi-mode motion feature learning.
Specifically, we employ a planning model based on structured scene representations as the teacher model, leveraging its diversified planning instances as multi-objective learning targets for the end-to-end model.
Moreover, we incorporate reinforcement learning to enhance the optimization of state-to-decision mappings, while utilizing generative modeling to construct planning-oriented instances, fostering intricate interactions within the latent space.
We validate our model on the nuScenes and NAVSIM datasets, achieving a 50\% reduction in collision rate and a 3-point improvement in closed-loop performance compared to the baseline model.
Code and model are publicly available at \href{https://github.com/YuruiAI/DistillDrive}{https://github.com/YuruiAI/DistillDrive}
\end{abstract}    
\section{Introduction}
\label{sec:intro}

\begin{figure}
    \centering
    \includegraphics[width=0.47\textwidth]{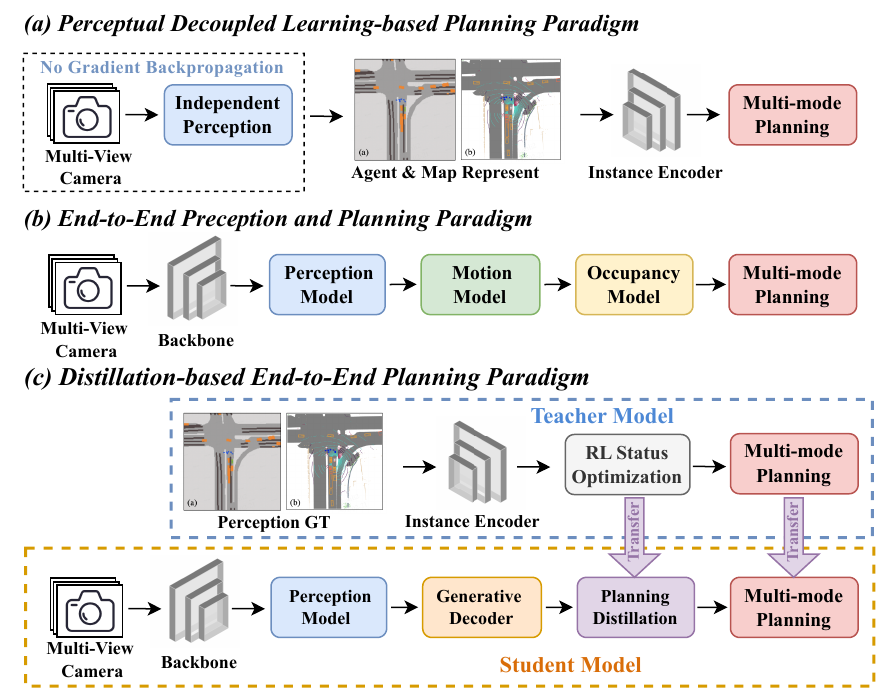}
    \caption{\textbf{Comparisons of different automatic driving planing paradigms}. (a) Perceptual decoupled learning-based planning model 
    \cite{pluto,diffusionplanner}. (b) End-to-end perception and planning model \cite{uniad, vad}. (c) The proposed distillation-based drive model.}
    \label{fig:fig1}
    \vspace{-0.1in}
\end{figure}

End-to-end autonomous driving \cite{uniad,vad,st-p3} has made significant progress in recent years, primarily driven by advancements in perception technologies \cite{sparse4d,LiSTM,CA-W3D} and imitation learning \cite{imitation}.
As show in \cref{fig:fig1} (b), it learns directly from complex sensor inputs to final planning and decision, eliminating the intermediate processes of data transfer and target characterization, thereby significantly reducing cascading errors \cite{uniad}.
However, in closed-loop experiments, the perceptually separated planning model in \cref{fig:fig1} (a) outperforms the end-to-end model, benefiting from contrastive learning \cite{pluto} and simulation experiments \cite{plankd1}.
Still, it faces a coupling barrier \cite{driveadapter} between perception and planning.

Unlike perceptual tasks with a unique solution, planning tasks have uncertain feasible solutions \cite{vadv2}, and relying on a single expert trajectory can limit the model's ability to learn diverse representations.
To address the issue of a limit planning goal, SparseDrive \cite{sparsedrive} incorporates unimodal planning features into multi-mode position embedding, enabling diverse planning.
Meanwhile, VADV2 \cite{vadv2} abstracts planning tasks as a probabilistic action distribution, using the interaction between planning vocabulary and scenario tokens to sample a single action. 
DiffusionDrive \cite{diffusiondrive} proposes a truncated diffusion policy with multi-mode anchors and the diffusion schedule, enabling the model to learn denoising from gaussian distribution.
While the above approach offers some planning diversity, it still has several issues:
(a) \textbf{Single-mode Learning}: Its target remains the expert trajectory from log playback, which lacks diverse supervision \cite{hydra-mdp} for multi-mode features and motion properties. 
(b) \textbf{Status Leakage}: The model suffers from over-reliance on ego status \cite{ego-mlp} and lack of state-to-decision space optimization.
(c) \textbf{Motion-guidance Deprivation}: The above approach lacks a planning-oriented feature modeling and fails to promote the instance interactions in a motion-guided manner.

To address the above issues, we use a multi-mode decoupled planning model as a teacher to supervise motion-guided instance interaction in the end-to-end model through knowledge distillation, presented in \cref{fig:fig1} (c).
Unlike existing methods \cite{Roach,driveadapter,hydra-mdp} that focus on simulation environments and lack supervision of latent-space multi-mode features, we effectively implement multi-mode imitation.
Meanwhile, we employ reinforcement learning to improve the comprehensive understanding of motion status and a generative model to boost interaction with motion distribution features from the expert's trajectory in latent space.

To effectively improve the planning performance of the end-to-end model and tackle the challenge of imitation learning in capturing the complexity of the planning space, we propose \textbf{DistillDrive}, an end-to-end multi-mode autonomous driving framework based on the distillation of an isomorphic hetero-source planning model.
Overall, we design a knowledge distillation architecture using decoupled planning models to efficiently supervise multi-mode planning learning in end-to-end models.
To clarify the role of the ego status, we use inverse reinforcement learning and Q-learning to enhance the construction of state-to-decision relationships.
Finally, motion-guided cross-domain feature interactions are enabled through generative modeling, enhancing the abstraction of instances to the planning space.

The main contributions can be summarized as follows:
\begin{itemize}
\item We propose a distillation  architecture for multi-mode instance supervision in end-to-end planning, tackling single-target imitation learning limitations.
\item We introduce reinforcement learning-based state optimization to enhance state-to-decision space understanding and mitigate ego motion state leakage.
\item To address missing motion-guided attributes, we use a generative model for distribution-wise interaction between expert trajectories and instance features.
\item We conduct open- and closed-loop planning experiments on the nuScenes and NAVSIM datasets, achieving a 50\% reduction in collision rate and a 3-point increase in both EP and PDMS over the baseline.
\end{itemize}
\section{Related Work}
\label{sec:releated work}

\noindent\textbf{End-to-End Planning.} 
Autonomous driving is rapidly advancing as end-to-end concepts \cite{e2ereview,e2ereview1} gain popularity, with works \cite{transfuser, st-p3} achieving impressive results by using Transformers for spatial-temporal feature learning in planning models.
Notably, UniAD \cite{uniad} first integrates detection, tracking, and mapping using attention mechanisms, achieving strong planning performance, while VAD \cite{vad} simplifies these steps, balancing accuracy and performance with vectorized representation.
Unlike previous methods modeling planning as continuous trajectory learning, studies \cite{vadv2, hydra-mdp} abstracted motion space into a probabilistic decision space.
Meanwhile, Hu \etal~ \cite{mile} and Zheng \etal~ \cite{genad} enhance planning by modeling motion space as a gaussian distribution for capturing latent  representations.
While SparseDrive\cite{sparsedrive} propose a sparse query-centric paradigm for end-to-end autonomous driving, achieving strong performance.
However, researches \cite{ad-mlp,ego-mlp} show that existing end-to-end planning models overly rely on ego status, lack of interaction with other agents.
To address this, we design an end-to-end model that enhances distribution interaction and utilizes expert trajectories for supervised instance-wise representation, providing motion prior.

\noindent\textbf{Knowledge Distillation.} 
Knowledge distillation (KD) allows a compact student model to mimic the behavior of a complex teacher model, inheriting its embedded knowledge \cite{msd,one}.
Hao \etal~ \cite{OFAKD} proposes a cross-architecture method that aligns intermediate features into a logits space to distill knowledge from heterogeneous models.
Simultaneously, approaches \cite{FASD, lanecmkt,SUMMER} enhance model expressiveness through multi-stage adaptive distillation and a dual-path mechanism.
In autopilot planning, Roach \cite{Roach} designed an RL-based expert to guide the IL agent in learning state space representations.
In PlanKD \cite{plankd1}, planning features are distilled, with trajectories and attention mechanisms extracting the feature center.
While Hydra-MDP  \cite{hydra-mdp} supervises multimodal trajectory generation through a rule-based teacher.
In contrast, our work tackles motion feature learning for cross-models via knowledge distillation, addressing the limitations of single-goal imitation learning.

\begin{figure*}[t] 
    \centering
    \includegraphics[width=0.95\textwidth]{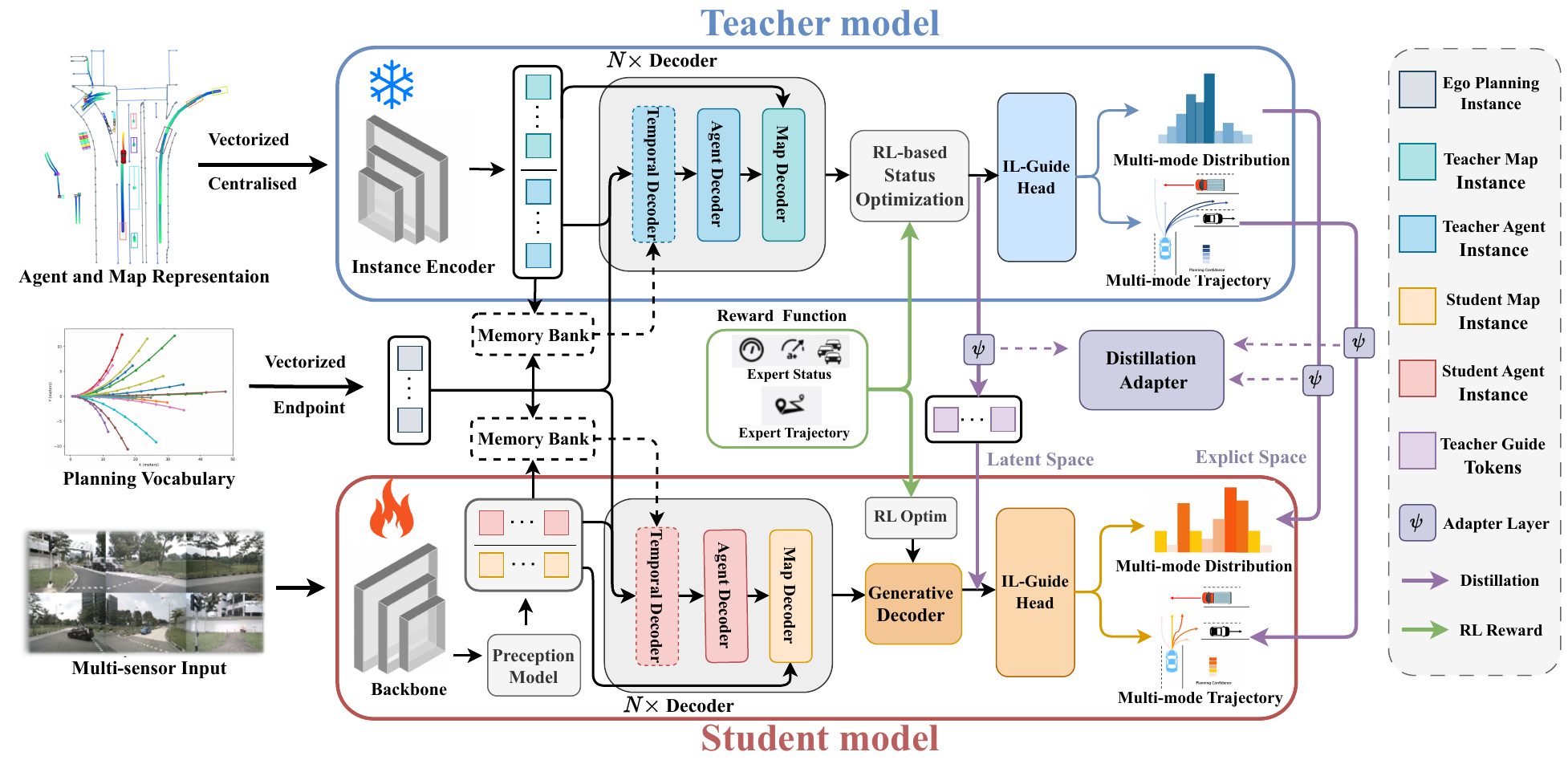}  
    \vspace{-0.1in}
    \caption{The Overview of our proposed DistillDrive. Initially, we train a teacher model with scene-structured annotation data, integrating reinforcement and imitation learning to enhance multi-mode planning. 
    Subsequently, we constructed an end-to-end student model and used a generative model to implement motion-oriented distribution interactions in latent space.
    Ultimately, multi-stage knowledge distillation and multi-mode supervision synergistically enhance the planning diversity and safety margins of autonomous driving models.}
    \vspace{-0.1in}
    \label{fig:fig2}
\end{figure*}

\noindent\textbf{Reinforcement Learning in Planning.} 
With the A3C agent \cite{A3C} applied in CARLA \cite{carla} using reinforcement learning, more studies are emerging in the field of autonomous driving.
Roach \cite{Roach} trains a reinforcement learning expert to translate bird's-eye view images into continuous low-level actions, setting a performance benchmark in CARLA.
RL methods have been applied to tasks like lane-keeping \cite{lanekeeping} and lane changing \cite{lanchange}, enabling policies to learn from rewards through closed-loop training.
Methods \cite{CQL} combine RL and IL to standardize Q-learning and prevent out-of-distribution value overestimation.
Inverse reinforcement learning infers cost functions from expert demonstrations, eliminating manual specification.
To better approximate the partition function, \cite{DrivingIRL, IRLPlanning} converts continuous behavioral modeling into a discrete setup, using a maximum entropy IRL architecture.
In practice \cite{INE}, actor goals are achieved by combining imitation learning and reinforcement learning, using the Soft Actor-Critic approach \cite{SAC} to alternately train critics and actors.
In our work, we combine Q-learning \cite{CQL} and IRL \cite{DrivingIRL} methods to improve the model's understanding of the state-to-decision space, enhancing both open- and closed-loop performance.

\section{Methodology}

\subsection{Overview}
As illustrated in \cref{fig:fig2}, we leverage a knowledge distillation-based planning model that improves multi-mode feature interaction through planning-based information.
In \cref{sec:teacher}, we present the teacher model's setup, including the encoder, planning decoder, IL-based head, and RL-based optimization.
Following this, we explore end-to-end planning models that leverage generative models and sparse feature representations in \cref{sec:student}.
Afterward, in \cref{sec:kd}, we focus on our distillation scheme, integrating multi-mode instance distillation to enhance feature learning.
Finally, we present the overall model training objectives in \cref{sec:target}, combining knowledge distillation and reinforcement learning to overcome the limitations of single-mode representation in imitation learning for planning tasks.

\subsection{IRL-based Teacher Model}
\label{sec:teacher}
The teacher model learns scene-to-planning relationships through an agent encoder, scene encoder, planning decoder, an imitation learning-based prediction head, and reinforcement learning-based state optimization.

\noindent\textbf{Agent Encoder.} 
For a dynamic target agent, its state at each moment \(t\) is characterized by the following information: position \(p_{i}^{t}\), orientation angle \(\theta_{i}^{t}\), velocity \(v_{i}^{t}\), dimensions \(b_{i}^{t}\) and binary mask information \(m_{i}^{t}\).
To enhance the characterization of the target's motion properties, we vectorize the features to correlate information across successive frames, modeling the target's dynamic characteristics: \(\bm{S}_{i}^{t} = \{p_{i}^{t}-p_{i}^{t-1},\theta_{i}^{t}-\theta_{i}^{t-1},v_{i}^{t}-v_{i}^{t-1},b_{i}^{t},m_{i}^{t}\}.\)
Then, an MLP is applied to encode the \( \bm{S} \in \mathbb{R}^{N_A \times T \times 9} \), resulting in \( \bm{E_A} \in \mathbb{R}^{N_A \times T \times D} \), where \( N_A \), \( T \), and \( D \) denote the number of agents, frames per agent, and dimension, respectively.

\( \bm{E_A} \) incorporates the agent's temporal features, enhancing through position encoding \( \bm{TE} \) based on sequence positions.
The global position embedding \( \bm{PE_A} \) is generated by encoding the position \(p_{i}^{t_{0}}\), providing rich spatial information and an encoder captures correlations among target temporal awareness via an attention mechanism:
\begin{equation}
  \bm{I_{A}} = Attn(Q(\bm{E_{A}}+\bm{TE}),K(\bm{E_{A}}+\bm{TE})),V(\bm{E_{A}})).
  \label{eq:2}
\end{equation}

\noindent\textbf{Scene Encoder.} 
Static targets (\eg, lane lines, road edges) help the model understand driving rules and areas, represented by a point set \( \bm{M} \in \mathbb{R}^{N_L \times N_P \times 2} \), where \( N_L \) is the number of lanes and \( N_P \) is the number of points per lane.
The lane center point \( \mathbf{c_{\text{m}}} \in \mathbb{R}^{N_L \times 2} \) serves as the position to encode \( \bm{PE_M} \), representing lane distribution.
The vectorized representation \( \mathbf{v_{\text{m}}} \), the orientation \( \mathbf{\theta_{\text{m}}} \), and deviation \( \mathbf{d_{\text{m}}} \) are derived to refine the lane center encoding.
Meanwhile, the labeling attribute \( \mathbf{l_{\text{m}}} \) for each lane is encoded through a weighting matrix \( \mathbf{\omega_{\text{m}}} \).
Finally, the lane lines are encoded using the PointNet \cite{pointnet} polyline encoder, resulting in an effective representation \( \bm{I_M} \in \mathbb{R}^{N_L \times D} \) as follows:
\begin{equation}
  \bm{I_{M}} = PointNet(Concat\{\mathbf{v_{m}},\mathbf{\theta_{m}},\mathbf{d_{m}}\}) + \mathbf{l_{m}} * \mathbf{\omega_{m}}.
  \label{eq:4}
\end{equation}

\noindent\textbf{Planning Decoder.}
As shown in \cref{fig:fig2}, we derive \( \bm{I_{E}}, \bm{PE_{E}} \in \mathbb{R}^{N_E \times D} \) from the planning vocabulary \cite{vadv2}, obtained by clustering trajectories based on vectorization and end-points, where \( N_E \) denotes the number of modes.
These multi-mode instances, along with features from the agent encoder, achieve cross-temporal aggregation through the memory bank \cite{sparsedrive} to enhance multi-frame agent and ego instances \( \bm{I'_{A/E}}\) and position embeddings \( \bm{PE'_{A/E}} \).

To model the mutual interactions among Ego, Agent, and Map, we employ attention mechanism for representation learning across heterogeneous features, capturing their interactions.
A temporal attention mechanism first captures associations between temporal features \( \bm{I'_{E/A}} \) to comprehend the target's evolving dynamics, while a self-attention further captures Ego-Agent interactions and analyzes their behavioral patterns.
Finally, cross-attention mechanism between the dynamic target \( \bm{I_{E/A}} \) and the static element \( \bm{I_M} \) enhances the agent's ability to abstract scenarios. Specific details are provided in the supplementary material.

\noindent\textbf{Imitation Learning-based Prediction Head.} 
After obtaining the multi-mode ego instance \( \bm{I_{E}}\), we use the MLP to predict trajectory regression \( \tau \in \mathbb{R}^{N_E \times T \times 2} \), classification \( s \in \mathbb{R}^{N_E \times 1} \), and ego-status \( \xi \in \mathbb{R}^{N_E \times 10} \) within a multi-task framework.
Both the regressed trajectories and motion status are supervised using L1 loss, yielding \( L_{\text{reg}} \) and \( L_{\text{status}} \), while the classification loss is enforced using KL divergence \cite{vadv2}, resulting in \( L_{\text{cls}} \) for multi-mode planning.
The details are provided in the appendix, where the imitation learning error \( L_{IL} \) is the sum of the three losses.

\noindent\textbf{Reinforcement Learning-based Status Optimization.} 
Ego status is learned from ego instances to prevent leakage and reduce state reliance \cite{ego-mlp}.
The planning solution space is diverse due to motion complexity, while imitation learning \cite{imitation}'s single objective limits model generality.
Thus, integrating reinforcement learning principles to optimize the policy space across statuses and behaviors is key to enhancing the model’s interaction and adaptability.

First, based on the predicted trajectory \({\hat{p}}_{traj}\) and status \(\hat{s}\), the per-point error, collision rate, and speed limit are calculated relative to the expert trajectory \({\bar{p}}_{traj}\) and status \(\bar{s}\).
The values are smoothed and normalized using a negative exponential function as the reward, with a learnable weight \(\omega_i\) applied to each reward to optimize the planning reward: \( r_t = \sum_{i=1}^{N_R} \omega_i e^{-x_i} \), where \(N_R\) is the number of rewards.
Unlike maximum entropy inverse reinforcement learning \cite{DrivingIRL}, which maximizes rewards via entropy, we refine weight coefficients through learning and designed reward rules for more intuitive computation.

Subsequently, Q-Learning \cite{dqn} is applied to map state values \(\hat{s}\) to the motion decision \(\bar{a}\), simplified to turn left, turn right, and go straight.
The expert's motion state \(\bar{s}\) is passed to the target network \(Q_T\) to learn behavioral decisions, with the decisions \(\bar{a}\) supervising the network, as shown in \cref{eq:6}.
Due to variability between \(\hat{s}\) and \(\bar{s}\), we avoid the target network updating and use separate supervision to clarify the association between state space and decisions.
We then compute the Q-value \( y_t \) for the scene using the target network, where \( \gamma \) is the reward decay coefficient:  
\begin{equation}
  y_{t} = r_{t} + \gamma  \max_{\bar{a}}Q_{T}(\bar{s}, \bar{a}).
  \label{eq:5}
\end{equation}

Finally, \( L_{RL} \) is computed as the difference between the main network’s prediction and the value \( y_t \), optimizing the model's state space representation for behavioral decisions:
\begin{equation}
  L_{RL} =  \sum_{i=1}^{\bar{a}} \| Q_{T}(\bar{s},\bar{a}_{i}) - \bar{a}_{i} \| +  \| Q_{M}(\hat{s}, \bar{a}_{max}) - y_t \|.
  \label{eq:6}
\end{equation}

\subsection{Motion-Guided Student Model}
\label{sec:student}

\noindent\textbf{Sparse Scene Representation.}
For the perceptual part of the model, we retain the SparseDrive \cite{sparsedrive} design, which uses fully sparse temporal representations.
The instance \(\bm{I_A}, \bm{I_M}\) remains the same structure as described in \cref{sec:teacher}.

\begin{figure}
    \centering
    \includegraphics[width=0.45\textwidth]{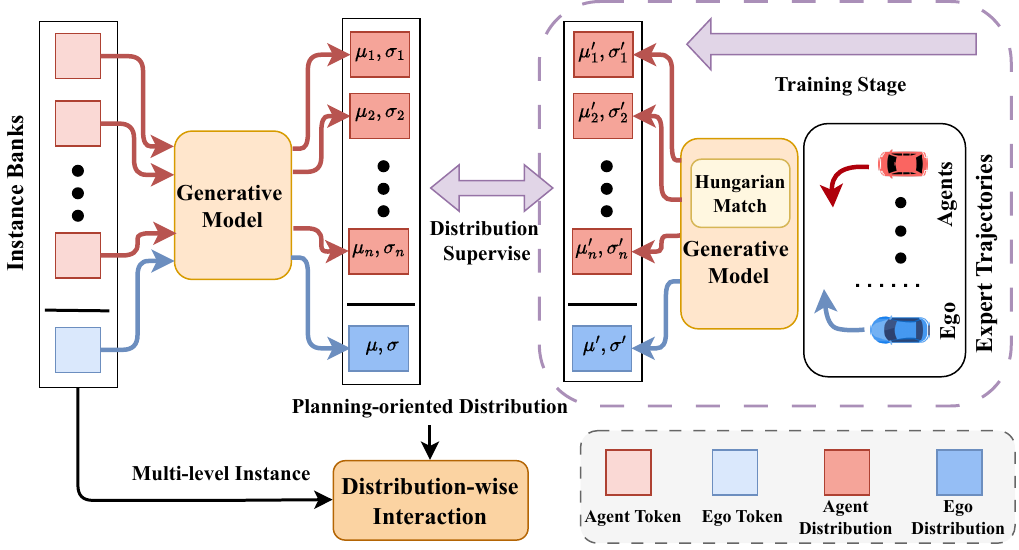}
    \caption{The Generative Models abstract expert trajectories, agent and ego feature as Gaussian distributions during training, using planning-oriented distribution to fuse instance tokens across multiple levels for distribution-wise interaction.}
    \vspace{-0.2in}
    \label{fig:fig3}
\end{figure}

\noindent\textbf{Planning-oriented Interactions in Generative Models.} 
To enhance the motion representation for Ego and Agent instances, we abstract \( \bm{I_{E/A}} \) and expert trajectories as Gaussian distributions, improving planning performance through distribution-wise interactions in the hidden space \( \mathbb{Z} \).
As shown in \cref{fig:fig3}, we map the expert trajectory \({\bar{p}}_{traj}\) to a Gaussian distribution in latent space \(\mathbb{Z}\) via an encoder, learning \(\mu_F,\sigma_F\) to model the distribution \(d(\mathbb{Z}|\bar{p}_{traj}) = N(\mu_{F}, \sigma_{F})\), capturing the prior of expert trajectories.

Similarly, for the instance \(\bm{I_E},\bm{I_A}\) of the Agent and Ego, we also use an encoder to map them to Gaussian space \(\mathbb{Z}\), obtaining \(\mu_I\) and \(\sigma_I\) to establish the distribution \(d(\mathbb{Z}|\bm{I_E}, \bm{I_A})= N(\mu_{I},\sigma_{I})\) for instance abstraction.

The training process uses \( N(\mu_{F}, \sigma_{F})\) for the interaction of motion features and supervises \(N(\mu_{I},\sigma_{I})\), while the inference process utilizes \( \mu_{I},\sigma_{I} \) to generate motion-guided features \(\bm{F_G} = \mu_{I} + n \times e^{\zeta \times \sigma_{I}}\).
Here, \(n\) denotes the Gaussian noise, \(\zeta\) represents the coefficient.
Then, \(\bm{I_E}, \bm{I_A}\) are concatenated and transformed dimensionally to obtain the multi-level instance \(\bm{I_G} \in \mathbb{R}^{L \times (N_E+N_A) \times D’}\), where \(L\) represents the level number.
Finally, the multi-scale instance \(\bm{I_G}\) serves as the hidden state, while the motion-guided feature \(\bm{F_G}\) is used as the input for distribution-level feature interaction via the GRU and MLP, as detailed in the appendix.
\begin{equation}
  (\bm{I_E},\bm{I_A}) = MLP(GRU(\bm{F_{G}},\bm{I_{G}})).
  \label{eq:7}
\end{equation}

For the distribution \( d(\mathbb{Z}|I_E, I_A) \) of the instance abstraction and the distribution \(d(\mathbb{Z}|\bar{p}_{traj})\) of the expert trajectory, we supervise the distributions using KL divergence.
The Ego distribution can be obtained directly, while the Agent target requires alignment using Hungarian algorithm \cite{hungarian}.
\begin{equation}
L_{DS} = \frac{\sigma_I - \sigma_F - 0.5 + e^{2 \times \sigma_F} + (\mu_F - \mu_I)^2}{2 \times e^{2 \times \sigma_I}}.
  \label{eq:13}
\end{equation}

\noindent\textbf{Planning Decoders and Prediction Heads.} 
As shown in \cref{fig:fig2}, the student model's decoder and motion attributes learning layer are aligned with the teacher model, preventing heterogeneity in subsequent distillation tasks.

\subsection{Knowledge Distillation}
\label{sec:kd}

In this section, we design a distillation architecture that leverages the multi-mode teacher model to supervise the student model through diversified imitation learning.

\noindent\textbf{Instance Distillation of Encoders.}
First, we apply L2 loss to align the encoder features of the teacher and student, \( \bm{F_{en}^{tc}} \) and \( \bm{F_{en}^{st}} \).
This method effectively aligns the shallow features of both models, ensuring consistency in the latent space, with the encoder distillation loss \( L_{en}^{KD} \) given by:
\begin{equation}
L_{en}^{KD}=\sum_{j=1}^{N_E} \left\| \bm{F_{en}^{st}} - \bm{F_{en}^{tc}} \right\|_2.
  \label{eq:14}
\end{equation}

\noindent\textbf{Instance Distillation of Decoders.} 
After the scene-wise and instance-wise decoders, we design a specialized adapter to accommodate the domain variability of the two models and address the heterogeneity between instances.
The MLP adapter \(\psi\) ensures isomorphism between the student and teacher decoder features, while the decoder distillation loss \( L_{de}^{KD} \) is given as follows:
\begin{equation}
L_{de}^{KD} = \sum_{i=1}^{N_E} \left\| \psi\left(\bm{F_{de}^{st}}\right) - \bm{F_{de}^{tc}} \right\|_2.
\end{equation}

\noindent\textbf{Motion Property Distillation.} 
Ultimately, to capture its probabilities across modes, we also design multi-mode distillation.
As shown below, we use KL divergence to supervise the multi-mode classifications \( p_{cls}^{st} \) and \( p_{cls}^{tc} \), encouraging the student to learn a distribution closer to the teacher.
\begin{equation}
L_{cls}^{KD} = \sum_{i=1}^{N_E} p_{cls}^{tc}(i) \log \left( \frac{p_{cls}^{tc}(i)}{p_{cls}^{st}(i)} \right).
\end{equation}

In summary, we design multi-stage distillation from the multi-mode teacher to the student model, addressing the challenge of a single learning objective.
By supervising features and probabilities, we enrich the model's representation of agent interactions, thereby enhancing planning performance.
The final multi-stage distillation loss is as follows:
\begin{equation}
L_{KD} = L_{en}^{KD} + L_{de}^{KD} +  L_{cls}^{KD}.\end{equation}

\subsection{Training Object}
\label{sec:target}
The learning objectives include the distillation loss \( L_{KD} \), distribution loss \( L_{DS} \), and reinforcement learning loss \( L_{RL} \). The imitation learning loss \( L_{IL} \) follows \cref{sec:teacher}, with total loss \( L_{T} \)  defined as follows, where \( \lambda_1, \lambda_2, \lambda_3 \) are the weight.
\begin{align}
L_{T} &= \lambda_{1} \times L_{KD} + \lambda_{2} \times L_{DS} + \lambda_{3} \times L_{RL}+L_{IL}.
  \label{eq:15}
\end{align}

\section{Experiments}

\begin{table*}[]
    \centering
    \resizebox{0.8\linewidth}{!}{
    \begin{tabular}{l|l|llll|llll|l}
    \hline
    \noalign{\hrule height 0.5pt}
    \multicolumn{1}{c|}{}                        &                         & \multicolumn{4}{c|}{L2(m)↓}                                                           & \multicolumn{4}{c|}{Collision(\%)↓}                                                    & \multicolumn{1}{c}{}                      \\
    \multicolumn{1}{c|}{\multirow{-2}{*}{Model}} & \multirow{-2}{*}{Input} & 1s            & 2s            & 3s            & \cellcolor[HTML]{EFEFEF}Avg.          & 1s            & 2s            & 3s            & \cellcolor[HTML]{EFEFEF}Avg.           & \multicolumn{1}{c}{\multirow{-2}{*}{FPS $\uparrow$}} \\ \hline
    FF \cite{ff}                                          & LiDAR                   & 0.55          & 1.20          & 2.54          & \cellcolor[HTML]{EFEFEF}1.43          & 0.06          & 0.17          & 1.07          & \cellcolor[HTML]{EFEFEF}0.43           & -                                         \\
    EO \cite{eo}                                           & LiDAR                   & 0.67          & 1.36          & 2.78          & \cellcolor[HTML]{EFEFEF}1.60          & 0.04          & 0.09          & 0.88          & \cellcolor[HTML]{EFEFEF}0.33           & -                                         \\ \hline
    ST-P3 \cite{st-p3}                                        & Camera                  & 1.33          & 2.11          & 2.90          & \cellcolor[HTML]{EFEFEF}2.11          & 0.23          & 0.63          & 1.27          & \cellcolor[HTML]{EFEFEF}0.71           & 1.6                                       \\
    UniAD \cite{uniad}                                        & Camera                  & 0.45          & 0.70          & 1.04          & \cellcolor[HTML]{EFEFEF}0.73          & 0.62          & 0.58          & 0.63          & \cellcolor[HTML]{EFEFEF}0.61           & 1.8                                       \\
    VAD \cite{vad}                                          & Camera                  & 0.41          & 0.70          & 1.05          & \cellcolor[HTML]{EFEFEF}0.72          & 0.03          & 0.19          & 0.43          & \cellcolor[HTML]{EFEFEF}0.21           & 4.5                                       \\
    SparseDrive \cite{sparsedrive}                                  & Camera                  & 0.31          & 0.60          & 1.00          & \cellcolor[HTML]{EFEFEF}0.63          & 0.01          & 0.08          & 0.30          & \cellcolor[HTML]{EFEFEF}0.13           & 6.5                                       \\ \hline
    DistillDrive (Teacher)                                     & Annotation                  & {0.27} & {0.51} & {0.82} & \cellcolor[HTML]{EFEFEF}{0.53} & {{0.01}}    & {{0.04}}    & {0.10} & \cellcolor[HTML]{EFEFEF}{0.05} & 31.6                                      \\
    DistillDrive (Student)                                     & Camera                  & \textbf{{0.28}}    & \textbf{0.54}    & \textbf{{0.83}}    & \cellcolor[HTML]{EFEFEF}\textbf{{0.57}}    & \textbf{0.00} & \textbf{0.03} & \textbf{{0.17}}    & \cellcolor[HTML]{EFEFEF}\textbf{{0.06}}    & 6.0                                       \\ \hline
    \noalign{\hrule height 0.5pt}
    \end{tabular}
    }
    \caption{Quantitative comparisons of planning performance using open-loop metrics on the nuScenes \cite{nuscenes} val dataset. SparseDrive results are replicated under the same settings, and the teacher model, based on annotations, is excluded from the performance comparisons.}
    \label{tab:tab1}
\end{table*}

\begin{table*}[]
    \centering
    \resizebox{0.8\linewidth}{!}{
    \begin{tabular}{l|cc|ccccc
    >{\columncolor[HTML]{EFEFEF}}c }
    \hline
    \noalign{\hrule height 0.5pt}
    Model          & Input & Backbone  & NC↑  & DAC↑ & TTC  & Comf.↑ & EP↑  & PDMS↑ \\ \hline
    Const Velocity & -     & -         & 69.0 & 57.8 & 58.0 & \textbf{100}    & 19.4 & 20.6  \\
    Ego Status MLP & -     & -         & 93.0 & 77.3 & 83.6 & \textbf{100}    & 62.8 & 65.6  \\ \hline
    UniAD \cite{uniad}          & C     & ResNet-34 & 97.8 & 91.9 & 92.2 & \textbf{100}    & 78.8 & 83.4  \\
    VADV2 \cite{vadv2}          & C     & ResNet-34 & 92.2 & 89.1 & 91.6 & \textbf{100}    & 76.0 & 80.9  \\
    Transfuser \cite{transfuser}     & C \& L  & ResNet-34 & 97.8 & 92.3 & 92.9 & \textbf{100}    & 78.6 & 83.5  \\ 
    Hydra-MDP \cite{hydra-mdp}     & C \& L  & ResNet-34 & 97.9 & 91.7 & 92.9 & \textbf{100}    & 77.6 & 83.0  \\ \hline
    DistillDrive (Teacher)       & GT     & -         & 97.5 & 96.0 & 92.8 & \textbf{100}    & 81.0 & 86.5  \\
    DistillDrive (Student)       & C \& L  & ResNet-34 & \textbf{98.1}    & \textbf{94.6}    & \textbf{93.6}    & \textbf{100}      & \textbf{81.0}    & \textbf{86.2}     \\ \hline
    \noalign{\hrule height 0.5pt}
    \end{tabular}
    }
    \caption{Quantitative comparison on planning-oriented NAVSIM \cite{NAVSIM} navtest split with closed-loop metrics. “C“ and "L" Denotes the use of camera and LiDAR sensor, While “GT” represents the use of ground truth annotations. Similarly, the teacher model is included only to demonstrate performance, not for actual performance comparison. The evaluation metrics are kept in default settings.
}
    \label{tab:tab2}
\end{table*}

\subsection{Dataset and Metrics}
To evaluate the performances of the end-to-end planning model, we conducted comparison and ablation experiments on the nuScenes dataset \cite{nuscenes}.
It includes ~1,000 driving scenes from Boston and Singapore, each lasting 20s, providing  camera, LiDAR, and annotation at 2Hz.
However, most scenes involve uniform motion, and the dataset mainly targets perception tasks, using open-loop evaluations for planning, making it ideal for performance assessment.

The NAVSIM dataset \cite{NAVSIM}, streamlined from nuPlan \cite{nuplan}, uses eight cameras for 360° coverage and fuses LiDAR point clouds from five sensors for enhanced perception.
The dataset provides high-quality annotations at 2Hz, including HD maps and object bounding boxes, ensuring reliable data for autonomous planning tasks.
Finally, the dataset focuses on challenging scenarios driven by dynamic changes in driving intentions, while excluding simpler cases like stationary or constant-speed driving, to better simulate autonomous planning in complex traffic environments.

\subsection{Experimental Settings }
To ensure a fair comparison, we use the same backbone and perception module as SparseDrive \cite{sparsedrive} on the nuScenes dataset, relying solely on panoramic images without additional LiDAR data.
We trained on 8 NVIDIA A800 GPUs with a total batch size of 48, using the AdamW optimizer.
The process involves training the teacher model for 30 epochs, followed by 100 epochs of perception pretraining. Then, the teacher and perception-pretrained models undergo 10 epochs of end-to-end distillation for the planning model.
On the NAVSIM dataset, we use the baseline settings of Transfuser \cite{transfuser}, with three cropped and scaled forward camera images (1024×256) and rasterized BEV LiDAR as input, without temporal training to ensure a fair comparison.
We trained using 8 NVIDIA A800 GPUs and the AdamW optimizer (learning rate \(1 \times 10^{-4}\)) for 100 epochs.
The process follows the same logic: first, the teacher model is trained separately, then used for end-to-end distillation of the student planning model. Where \(\lambda_1, \lambda_2, \lambda_3, \gamma, \zeta\) are 0.5, 1, 1, 0.95, and 0.5.

\subsection{Results and Analysis}

\noindent\textbf{Planning Performance on nuScenes Dataset.}
We validate the planning performance of the proposed DistillDrive on the nuScenes dataset, as shown in \cref{tab:tab1}.
The teacher model excels across all models, validating the effectiveness of reinforcement learning for state-space optimization, but it encounters coupling barriers \cite{driveadapter} in real-world.
Therefore, we propose a knowledge distillation scheme to enhance end-to-end multi-mode planning learning and validate our model’s performance in the table.
It outperforms both LiDAR-based \cite{ff,eo} and camera-based models \cite{st-p3,uniad,vad}, achieving a 50\% reduction in collision rate and a 10\% decrease in L2 error compared to SparseDrive. This demonstrates that diverse imitation learning effectively enhances the distinction between mode instances.

\noindent\textbf{Planning Performance on NAVSIM Dataset.}
\cref{tab:tab1} focuses on open-loop performance with assessment limitations, while \cref{tab:tab2} offers a comprehensive evaluation using NAVSIM’s data-driven, non-reactive simulation.
Results show the teacher model excels in performance, confirming the validity of its design.
Through effective multi-mode instance imitation, our end-to-end student model outperforms the Transfuser \cite{transfuser} by 2.5\% on PDMS, with significant gains in DAC and EP.
DistillDrive also surpass rule-based distillation \cite{hydra-mdp} with significant performance gains.
It even outperforms the teacher model on NC and TTC, demonstrating that multi-mode distillation not only transfers the teacher's knowledge but also enables significant breakthroughs.

\noindent\begin{table}
    \centering
    \renewcommand{\arraystretch}{1.1}
    \setlength{\tabcolsep}{1.0mm}
    \resizebox{1.0\linewidth}{!}{
    \begin{tabular}{@{}lccccccc@{}}
        \hline
        \noalign{\hrule height 0.5pt}
        \textbf{Model} & \textbf{NDS$\uparrow$} & \textbf{mAP$\uparrow$} & \textbf{mATE$\downarrow$} & \textbf{mASE$\downarrow$} & \textbf{mAOE$\downarrow$} & \textbf{mAVE$\downarrow$} & \textbf{mAAE$\downarrow$} \\
        \hline
        UniAD & 49.8 & 0.38 & 0.684 & 0.277 & 0.383 & 0.381 & 0.192 \\
        SparseDrive & 52.5 & 0.418 & \textbf{0.565} & \textbf{0.275} & 0.552 & \textbf{0.261} & 0.190 \\
        DistillDrive   & \textbf{53.0} & \textbf{0.420} & 0.568 & \textbf{0.275} & \textbf{0.539} & {0.265} & \textbf{0.1766} \\
        \hline
        \noalign{\hrule height 0.5pt}
    \end{tabular}
    }
    \vspace{-0.1in}
    \caption{Quantitative Comparisons with end-to-end methods in detection performance on the nuScenes val dataset.}
    \label{tab:tab3}

\end{table}

\begin{table}[]
    \renewcommand{\arraystretch}{1.1}
    \resizebox{1.0\linewidth}{!}{
    \begin{tabular}{l|llll|llll}
    \hline
    \noalign{\hrule height 0.5pt}
    Model        & AMOTA$\uparrow$          & AMOTP$\downarrow$          & Recall$\downarrow$          & IDS$\downarrow$          & ${AP_{c}}$$\uparrow$   & ${AP_{d}}$$\uparrow$   & ${AP_{b}}$$\uparrow$    & mAP$\uparrow$           \\ \hline
    UniAD        & 0.359          & 0.1320         & 04.67           & 906          & -             & -             & -             & -             \\
    VAD          & -              & -              & -               & -            & 40.6          & 51.5          & 50.6          & 47.6          \\
    SparseDrive  & \textbf{0.385}          & 1.254          & \textbf{0.499}           & 886          & \textbf{49.9}          & 57.0          & \textbf{58.4}          & \textbf{55.0}          \\
    DistillDrive & {0.371} & \textbf{1.252} & {0.500} & \textbf{767} & {49.5} & \textbf{57.3} & {58.3} & \textbf{55.0} \\ \hline
    \noalign{\hrule height 0.5pt}
    \end{tabular}
    }
    \vspace{-0.1in}
    \caption{Comparison of multi-target tracking and mapping performance on the nuScenes dataset. Here, \( AP_{c} \), \( AP_{d} \), and \( AP_{b} \) represent the AP of crossing, divider, and boundary, respectively.}
    \vspace{-0.1in}
    \label{tab:tab4}
\end{table}

\begin{figure*}[t] 
    \centering
    \includegraphics[width=0.95\textwidth]
    {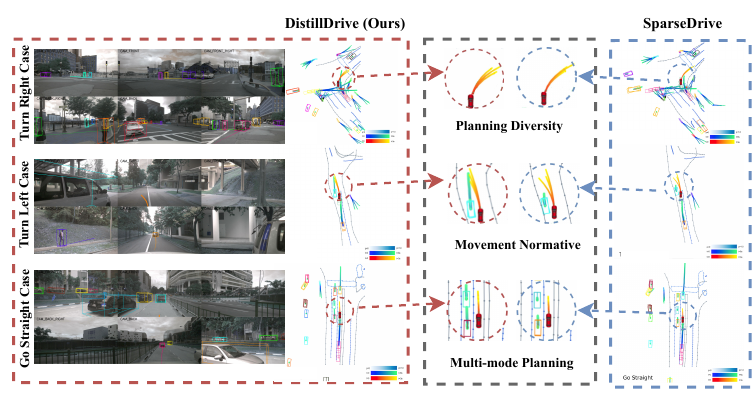}  
    \vspace{-0.05in}
    \caption{Qualitative visualization with end-to-end methods in planning performance on the nuScenes val dataset. We conduct a comprehensive comparison with SparseDrive, focusing primarily on their planning performance in the bird's-eye view(BEV) space.
    The first and third rows show more diverse plans, while the second row exhibits better kinematic alignment under the same modality.}
    \vspace{-0.1in}
    \label{fig:fig4}
\end{figure*}

\begin{figure}
    \centering
    \includegraphics[width=0.40\textwidth]{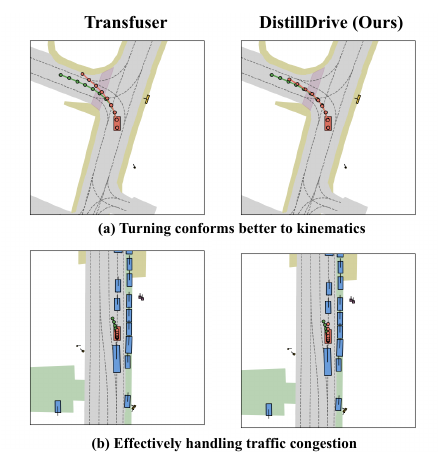}
    \caption{Qualitative visualization of the planning performance on the NAVSIM navtest split, where the red line represents the predicted trajectory and the green line represents expert trajectory.}
    \vspace{-0.2in}
    \label{fig:fig5}
\end{figure}

\noindent\textbf{Perception Performance on nuScenes Dataset.}
To evaluate our planning model, \cref{tab:tab3} and \cref{tab:tab4} sequentially validate its detection, tracking, and mapping performance. Since no additional perception module was designed, the overall performance is similar to SparseDrive. However, we achieved improvements in metrics like IDS, thanks to the contribution of generative modeling.

\noindent\textbf{Qualitative Visualization.}
\cref{fig:fig4} visualizes the planning model's performance on the nuScenes dataset across three scenarios: right turn, left turn, and straight.
The right-turn case shows greater trajectory diversity, highlighting the role of multi-mode instance distillation.
In the left-turn case, our model produces smoother, more natural trajectories due to effective RL-based status optimization.
The straight case shows a richer set of trajectory candidates, demonstrating enhanced planning capabilities.
In \cref{fig:fig5}, we validate the model's planning performance on NAVSIM dataset. In the first case, our model’s trajectory closely matches the expert, demonstrating better completion. In the second case, while Transfuser fails due to braking, our model successfully overtakes, handling congestion effectively.

\begin{figure}
    \centering
    \includegraphics[width=0.38\textwidth]{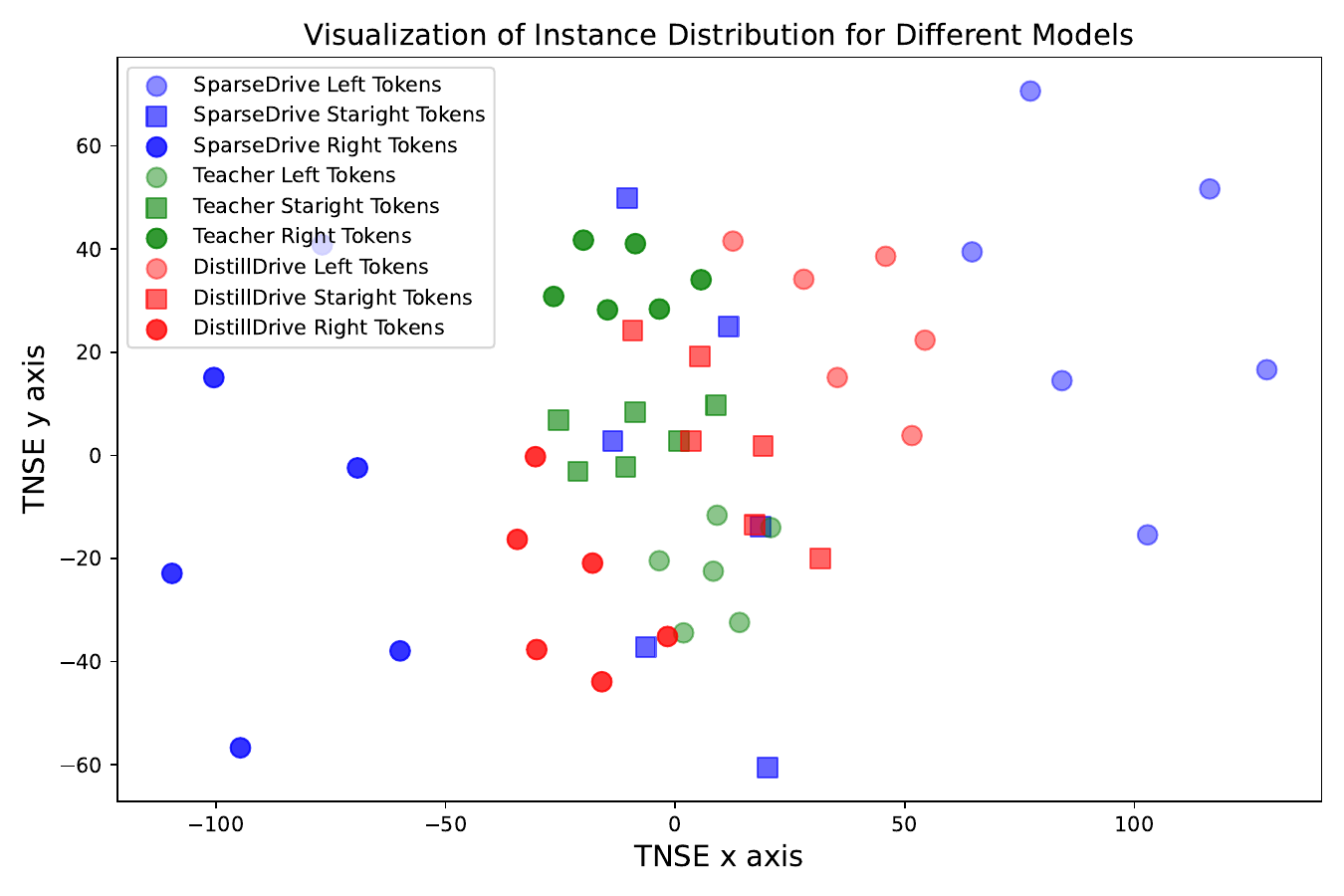}
    \caption{Visualization of multi-mode instances using t-SNE across SparseDrive, the teacher model, and DistillDrive.}
    \label{fig:fig6}
\end{figure}

To analyze the impact of multi-mode instance imitation on planning tokens, we use t-SNE to project tokens into two-dimensional space and compare SparseDrive, the teacher model, and DistillDrive in \cref{fig:fig6}.
DistillDrive compresses spatial representations while preserving the original model \cite{sparsedrive} distribution, learning intermediate representations through isomorphic model distillation, which demonstrates the feasibility of diverse instance imitation.

\subsection{Ablation Studies}


\noindent\textbf{Effect of Reinforcement Learning Optimization.}
In \cref{tab:tab6}, we apply reinforcement learning to refine the model, enhancing its comprehension of status and decision space. 
Comparing the second to fourth lines indicates that appropriate supervision and rewards enhance trajectory planning, while action selection in the fifth line refines performance. 
Finally, the proposed linear weighted reinforcement learning dynamically integrates various reward values.
This approach achieves a 20\% reduction in collision rate and a 6\% decrease in alignment L2 error, demonstrating its effectiveness in optimizing planning accuracy and safety.


\begin{table}[]
    \renewcommand{\arraystretch}{1.1}
    \resizebox{1.0\linewidth}{!}{
    \begin{tabular}{llll|ll}
    \hline
    \noalign{\hrule height 0.5pt}
    ${L_{RL}}$   & Reward       & Select       & IRL          & Avg Collision(\%) & Avg L2(m) \\ \hline
    \ding{55}   & \ding{55}  & \ding{55}  & \ding{55} & 0.066             & 0.5716    \\
    \ding{51} & \ding{55}  & \ding{55}  & \ding{55}  & 0.057             & 0.5777    \\
    \ding{51} & \ding{51} & \ding{55}  &  \ding{55}  & 0.063             & 0.5653    \\
    \ding{51} & \ding{51} & \ding{51} & \ding{55}   & 0.059              & 0.5522    \\
    \ding{51} & \ding{51} & \ding{51} & \ding{51} & \bfseries 0.053 \textcolor{
    Green}{\lowercase{-20\%}}             & \bfseries 0.5364 \textcolor{
    Green}{\lowercase{-6\%}}   \\ \hline
    \noalign{\hrule height 0.5pt}
    \end{tabular}
    }
    \caption{Ablation study on reinforcement learning in the nuScenes val, where \( L_{RL} \), Reward, Select and IRL represent loss, reward, state-space allocation and inverse reinforcement learning.}
    \vspace{-0.1in}
    \label{tab:tab6}
\end{table}

\noindent\textbf{Ablation Study of DistillDrive Modules.}
To validate the design's effectiveness, we assess the core modules of the end-to-end planning model in \cref{tab:tab7}. 
In rows second to fifth, individual evaluations reveal that Knowledge Distillation (KD) significantly enhances expressiveness via multi-mode supervision, while the generative model benefits from distribution-wise interaction for motion learning. However, Reinforcement Learning (RL) faces challenges due to limited state space representation. 
Meanwhile, result show that combining the components significantly improves model performance.
KD as core elements, enhance multi-mode motion imitation, while RL, through state-to-decision supervision, boosts generalization to out-of-distribution data, improving decision stability and robustness.
Further closed-loop experiment results are included in the appendix.

\noindent\textbf{Ablation Study on Generative Model.}
Analyzing \cref{tab:tab8} we observe that using Hungarian matching to align motion features encoded in the expert trajectory with those in the image decoding significantly improves planning and prediction performance.
While adjusting the $\zeta$ parameter has a minor effect, setting $\zeta$ to 0.5 yields consistently high performance.
Lastly, incorporating the autoregressive strategy negatively impacts performance by the difficulty of training.

\noindent\textbf{Effect of Knowledge Distillation.}
In \cref{tab:tab9}, we validate our knowledge distillation approach and demonstrate that the gains from distillation are more pronounced in the encoders and decoders, thereby facilitating diverse instance imitation learning.
The benefits of classification distillation are limited because the core planning functionality relies on regression heads, rendering classification supervision minimally impact on planning performance.
Additional details on regression distillation are provided in the appendix

\begin{table}[]
    \renewcommand{\arraystretch}{1.1}
    \resizebox{1.0\linewidth}{!}{
    \begin{tabular}{lll|ll}
    \hline
    \noalign{\hrule height 0.5pt}
    RL           & Distillation           & Generative          & Avg Collision(\%) & Avg L2(m) \\ \hline
    \ding{55}  & \ding{55}  & \ding{55} & 0.132             & 0.6398    \\
    \ding{51} & \ding{55} & \ding{55} & 0.137             & 0.6715    \\
    \ding{55} & \ding{51} & \ding{55} & 0.092             & 0.5851    \\
    \ding{55} & \ding{55} & \ding{51} & 0.107             & 0.5966    \\ \hline
    \ding{51} & \ding{51} & \ding{55} & 0.094             & 0.5915    \\
    \ding{55} & \ding{51} & \ding{51} & 0.101             & 0.5846    \\
    \ding{51} & \ding{55} & \ding{51} & 0.075             & 0.5968    \\
    \ding{51} & \ding{51} & \ding{51} & \bfseries 0.067 \textcolor{Green}{\lowercase{-50\%}}            & \bfseries 0.5735 \textcolor{Green}{\lowercase{-10\%}}    \\ \hline
    \noalign{\hrule height 0.5pt}
    \end{tabular}
    }
    \caption{Ablation study of designed modules on the nuScenes val.}
    \label{tab:tab7}
\end{table}

\begin{table}[]
    \renewcommand{\arraystretch}{1.1}
    \resizebox{1.0\linewidth}{!}{
    \begin{tabular}{lll|llll|lll}
    \hline
    \noalign{\hrule height 0.5pt}
    \multirow{2}{*}{$\zeta$} & \multirow{2}{*}{HM} & \multirow{2}{*}{AR} & \multicolumn{4}{c|}{Prediction}                                   & \multicolumn{3}{c}{Planning}                     \\
                        &                     &                     & EPA            & ADE            & FDE            & MR             & Collision      & ADE            & FDE            \\ \hline
    0.5  & \ding{55}   & \ding{55}  & 0.484          & 0.622          & 0.981          & 0.134          & 0.125          & 0.614          & 0.963          \\
    0.5  & \ding{51}        & \ding{55} & \textbf{0.539} & \textbf{0.422} & \textbf{0.538} & \textbf{0.045} & \underline{0.093}  & \textbf{0.583} & \textbf{0.914} \\
    1.0  & \ding{51}        &  \ding{55} & \underline{0.538}  & \underline{0.427}          & \underline{0.547}  & \underline{0.046}  & \textbf{0.091} & \underline{0.585}          & \underline{0.917} \\
    0.5  & \ding{51}        & \ding{51}        & 0.533          & 0.445          & 0.582          & 0.051          & 0.151          & 0.566          & 0.916          \\ \hline
    \noalign{\hrule height 0.5pt}
    \end{tabular}
    }
    \caption{Ablation study of prediction and planning performance with generative model on the nuScenes val set, where HM and AR denote Hungarian Matching and autoregressive strategy.}
    \label{tab:tab8}
\end{table}

\begin{table}[]
    \renewcommand{\arraystretch}{1.2}
    \resizebox{1.0\linewidth}{!}{
    \begin{tabular}{lll|ll}
    \hline
    \noalign{\hrule height 0.5pt}
    {$L_{en}^{KD}$} & {$L_{de}^{KD}$} & {$L_{cls}^{KD}$} & Avg Collision(\%) & Avg L2(m)       \\ \hline
    \ding{51}  & \ding{55} & \ding{55} & \textbf{0.084}    & 0.5884          \\
    \ding{55}& \ding{51}  & \ding{55} & 0.092             & \underline{0.5851}    \\
    \ding{55} & \ding{55} & \ding{51}   & 0.129             & 0.5857          \\ \hline
    \ding{51}  & \ding{55} & \ding{51}   & 0.096             & 0.5918          \\
    \ding{55}  & \ding{51}  & \ding{51}   & 0.102             & 0.6010          \\
    \ding{51}  & \ding{51}  & \ding{55}  & \underline{ 0.087}       & \textbf{0.5769} \\ \hline
    \noalign{\hrule height 0.5pt}
    \end{tabular}
    }
    \caption{Ablation study of distillation methods on nuScenes val.}
    \label{tab:tab9}
\end{table}

\section{Conclusion and Future Work}

To enhance the planning performance of End-to-end models, we implement a diversified instance imitation learning architecture to supervise the learning of multi-mode features.
Reinforcement learning and generative modeling enhance motion-guided feature interactions, leading to significant improvements on the nuScenes and NAVSIM datasets.

In future work, we aim to integrate world models with language models to enhance planning performance.
More effective reinforcement learning methods will be employed to better understand the relationship between the semantic geometric space of the scene and the decision planning space, enhancing the model closed-loop performance.
\section*{\textbf{Acknowledgements}}
This work is supported by National Natural Science Foundation of China (62333005), Shanghai International Science and Technology Cooperation Project (24510714000) and Innovation Program of Shanghai Municipal Education Commission (2021-01-07-00-02-E00105).
{
    \small
    \bibliographystyle{ieeenat_fullname}
    \bibliography{main}
}

\clearpage
\setcounter{page}{1}
\maketitlesupplementary

\section{Supplementary Methodology}
In this section, we present additional content that is not covered in detail in the paper to further support the theoretical framework of the argumentative article.

\subsection{IRL-based Teacher Model}
\noindent\textbf{Position Embedding.} 
For the temporal feature \( \bm{I_A} \), we apply temporal positional encoding \( \bm{TE} \), indexed by its sequence length, as described by the following equation.
\begin{align}
pos(i) = \begin{cases} 
\sin\left(\frac{pos}{10000^{\frac{2i}{D}}}\right),  & \text{if } i \text{ is even}, \\
\cos\left(\frac{pos}{10000^{\frac{2i}{D}}}\right).  & \text{if } i \text{ is odd}.
\end{cases}
\label{eqs:eq1}
\end{align}

For the global position encoding \( \bm{PE_{A/M}} \), the initial position \( pos_i^{t_0} \) is mapped to a higher-dimensional space in the same manner as \cref{eqs:eq1}, by passing through a linear layer and activation function to provide a global position prior.

\noindent\textbf{Map Representation.} 
For the map's feature encoding, \( \mathbf{c_m} \in \mathbb{R}^{N_L \times 2}\) denotes the location point at the indexed \( {N_P}/{2} \) position in the subset \( \bm{M}  \in \mathbb{R}^{N_L \times N_P \times 2} \) of lane lines.
The vectorized representation \( \mathbf{v_{\text{m}}} \) and deviation \( \mathbf{d_{\text{m}}} \) are calculated using the following formula, while the orientation \( \mathbf{\theta_{\text{m}}} \) is obtained by calculating the inverse tangent of \( v_{\text{m}} \).
\begin{equation}
  v^{i}_{j}, d_{j}^{i} = (\bm{M}^{i}_{j}-\bm{M}^{i-1}_{j}),  (\bm{M}^{i}_{j}-\bm{c_{m}}),
  \label{eqs:eq2}
\end{equation}
where \( j \) denotes the \( j \)-th lane line, \( i \) denotes the \( i \)-th point, and \( \bm{M}_{j}^{i} \) is a 2D coordinate in the lane line object.

\noindent\textbf{Ego Motion Encoder.}
As described in the main paper, we obtain \( \bm{I_E} \) and \( \bm{PE_E} \) by vectorizing and extracting endpoints from cluster center trajectories.
Vectorization, achieved by solving trajectory differences (similar to the operation of \( p_{i}^{t} \) in the Agent Encoder), enables us to abstract the motion behavior of clustered trajectories in motion space. 
It is then mapped to the implicit space via \cref{eqs:eq1} and further encoded by a learnable network layer to obtain the implicit embedding \( \bm{I_E} \).
For \( \bm{PE_E} \), we use its endpoints to represent the final target points across different modes and map them to a unified high-dimensional representation, similar to positional encoding.

\noindent\textbf{Temporal Decoder.}
All decoders take as input the concatenated Ego and Agent features from the current frame, denoted as $\bm{I} \in \mathbb{R}^{(N_{E}+N_{A}) \times 1 \times D}$, serving as the query. The corresponding position embedding is represented as $\bm{PE} \in \mathbb{R}^{(N_{E}+N_{A}) \times 1 \times D}$, where $N_E$, $N_A$, and $D$ denote the number of ego instances, agent instances, and feature dimensions, respectively.
In this decoder, the sequence length is represented by the time course, and the spatial-temporal correlation of each agent over time is captured through the cross attention mechanism.
For temporal features, its multi-frame information \( \bm{I_{T}}, \bm{PE_{T}} \in \mathbb{R}^{(N_{E}+N_{A}) \times T \times D} \) for each agent has been obtained through memory bank.
\begin{equation}
  \bm{I} = Attn(Q(\bm{I}+\bm{PE}),K(\bm{I_{T}}+\bm{PE_{T}}),V(\bm{I_{T}})).
  \label{eqs:3}
\end{equation}

\noindent\textbf{Agent Decoder.}
In the Agent Decoder, we focus on the interaction between Ego and Agent to achieve cross-target feature querying. Therefore, we use the number of targets \( (N_E + N_A) \) as the sequence length and enable feature interaction through the following self-attention mechanism.
\begin{equation}
  \bm{I} = Attn(Q(\bm{I}+\bm{PE}),K(\bm{I}+\bm{PE}),V(\bm{I})).
  \label{eqs:4}
\end{equation}

\noindent\textbf{Map Decoder.}
To better understand the guiding role of static scenes in the planning process, the map decoder is designed to interact with the Ego and Agent instances \(\bm{I} \).
It uses the Map instance \( \bm{I_M} \) and the lane center position embedding \( \bm{PE_M} \) as the key and value for the decoder.
\begin{equation}
  \bm{I} = Attn(Q(\bm{I}+\bm{PE}),K(\bm{I_M}+\bm{PE_M}),V(\bm{I_M})).
  \label{eqs:5}
\end{equation}

\noindent\textbf{Imitation Learning Loss.}
The planning model predicts the trajectory \({\hat{p}}_{traj}\) directly using the expert trajectory \({\bar{p}}_{traj}\) constraints to obtain \( L_{\text{reg}} \).
In our setup, the objective is to learn the displacement between frames, effectively avoiding the issue of excessive variance in the regression values:
\begin{equation}
  L_{reg} = \sum_{i=1}^{T} \| {\hat{p}}_{traj} - {\bar{p}}_{traj} \|.
  \label{eqs:6}
\end{equation}

Similarly, the predicted self-vehicle status  \(\hat{s} \) are constrained using the expert driver’s status  \(\bar{s}\) (\(N\) variables as acceleration, angular velocity, speed, etc.).
\begin{equation}
  L_{status} = \sum_{i=1}^{N} \| {\hat{s}} - {\bar{s}} \|.
  \label{eqs:7}
\end{equation}

The classification loss \(L_{cls}\) leverages expert probability distribution supervision, enabling the model to capture richer information for multi-mode planning.
In our setup, the clustered trajectory closest to the expert trajectory is assigned the highest confidence (0.8), while the remaining nearest neighbors use soft labels to promote diversity in model learning, enabling multi-mode planning.
Here, the manually assigned expert probability distribution is denoted as \({\bar{p}}_{\text{cls}} \), while the predicted probability is \( {\hat{p}}_{\text{cls}} \), enabling diverse planning representations.
We then apply KL divergence \cite{vadv2} to effectively supervise the predictive distribution.
\begin{equation}
L_{cls} = \sum_{i=1}^{M_E} {\bar{p}}_{\text{cls}}(i) \log \left( \frac{{\bar{p}}_{\text{cls}}(i)}{{\hat{p}}_{\text{cls}}(i)} \right).
\label{eqs:8}
\end{equation}

The final imitation learning error is shown below, with its loss weights following the base setting of SparseDrive \cite{sparsedrive}, and no further ablation analysis provided.
\begin{align}
L_{IL} &= 2 \times L_{reg} + 3 \times L_{status} + 0.5 \times L_{cls}.
  \label{eqs:9}
\end{align}

\noindent\textbf{Reward Function.}
First, the state error \( e_{st} \) is defined similarly to \cref{eqs:7} to measure the difference in state estimates between the predicted state and the expert trajectory.
Similarly, the trajectory mean error \( e^{mean}_{traj} \) is calculated as in \cref{eqs:6}, while the trajectory start error and end point error \( e^{start}_{traj} \), \( e^{end}_{traj} \) are calculated only at specific points to measure the closeness of the model's prediction to the expert.

Next, the speed predicted \( \hat{s}_{vx} \) is monitored with thresholds to prevent obvious out-of-bounds behavior, as follows.
\begin{align}
e_{speed} = \begin{cases} 
1, & \text{if } 0 < \hat{s}_{vx} < 20, \\
0. & \text{otherwise}.
\end{cases}
\label{eqs:10}
\end{align}

Additionally, for state predictions \(\hat{s}\) and planning trajectories \({\hat{p}}_{traj}\), we evaluate their consistency error to encourage the model to establish associations between learned states and planning as following:
\begin{align}
e_{consist} &= \hat{s}^{t_{1}}_{vx} \times \Delta_{t} - {\hat{p}}_{traj}^{t_{1}}.
  \label{eqs:11}
\end{align}

Meanwhile, for the collision error \( e_{collision} \), the predicted trajectory and the real target are used to compute a value of 0 for collision and 1 for safety.

Finally, the negative exponent is used as a reward value, as described in the main text, and reward-weighted summation is implemented using linearly weighted reinforcement learning to enhance the reward representation of the scene.
\begin{align}
 r_t = \sum_{i=1}^{N} \omega_i e^{-x_i} .
  \label{eqs:12}
\end{align}

Unlike the experience replay strategy in DQN \cite{dqn}, our multi-batch setup enables a unified representation across multiple scenarios.
The proposed Target Network Update in DQN addresses target value fluctuations during training using a delayed update technique.
However, in our setup, the heterogeneous state values of the inputs to the main and target networks cause the delayed update policy to fail.
Therefore, we use the actual decision behavior \(\bar{a}\) for the individually supervised target network.
\begin{equation}
  L_{Target} =  \sum_{i=1}^{\bar{a}} \| Q_{T}(\bar{s},\bar{a}_{i}) - \bar{a}_{i} \| .
  \label{eqs:13}
\end{equation}

\subsection{Motion-Guided Student Model}
\noindent\textbf{Generative Encoder.}
Separate encoders are used to map expert trajectories and agent instances into Gaussian-distributed feature spaces.
A multi-layer 1D convolution with ReLU activation is applied to each token to parameterize the agent’s target distribution, enabling distribution-level interaction through fusion and supervision, as described in the main text.

\subsection{Knowledge Distillation}
\noindent\textbf{Adapter.}
As described in the paper, one set of agent and map features in the teacher and student models originates from the vectorized representation of annotation results, while the other is extracted from image features using deformable attention \cite{deformable} to capture key information.
To address feature heterogeneity after the decoder, we employ an adapter \(\psi\) for alignment.
\begin{equation}
  \psi =  Linear(ReLU(Linear)) .
  \label{eqs:14}
\end{equation}

\noindent\textbf{Motion Property Distillation.}
The categorized kinematic attribute distillation loss was introduced in the main text; however, its performance was found to be limited in certain experiments.
To address this, an additional regression-based attribute distillation loss was incorporated into the section.
Specifically, the strategy is similar to \cref{eqs:6} in that it uses the trajectory predictions \( p_{reg}^{tc} \) from the teacher model to supervise the student model trajectory \( p_{reg}^{st} \).
\begin{equation}
  L_{reg}^{KD} = \sum_{i=1}^{T} \| p_{reg}^{st} - p_{reg}^{tc}\|.
  \label{eqs:15}
\end{equation}

\section{Supplementary Experiments}
\label{sec:formatting}
Here, we provide a complementary account of ablation  experiments and visual analyses to illustrate the improvements in the model's performance.

\subsection{Ablation Studies}

\noindent\textbf{Ablation Study in Closed-Loop Evaluation}
While the paper focused on ablation studies with nuScenes, we have since conducted further experiments on the NAVSIM dataset \cite{NAVSIM} to assess the effectiveness of each module.
As shown in \cref{tabs:tab4}, distillation with the teacher model improves EP and PDMS, leading to more diverse and complete trajectories.
The generative model enhances planning by interacting with the underlying motion distribution.
However, since Transfuser is inherently trained without temporal modeling and relies on explicit ego status inputs, it conflicts with the implicit ego status representation and temporal learning in RL optimization, resulting in no significant performance improvement.

\begin{figure}
    \centering
    \includegraphics[width=0.45\textwidth]{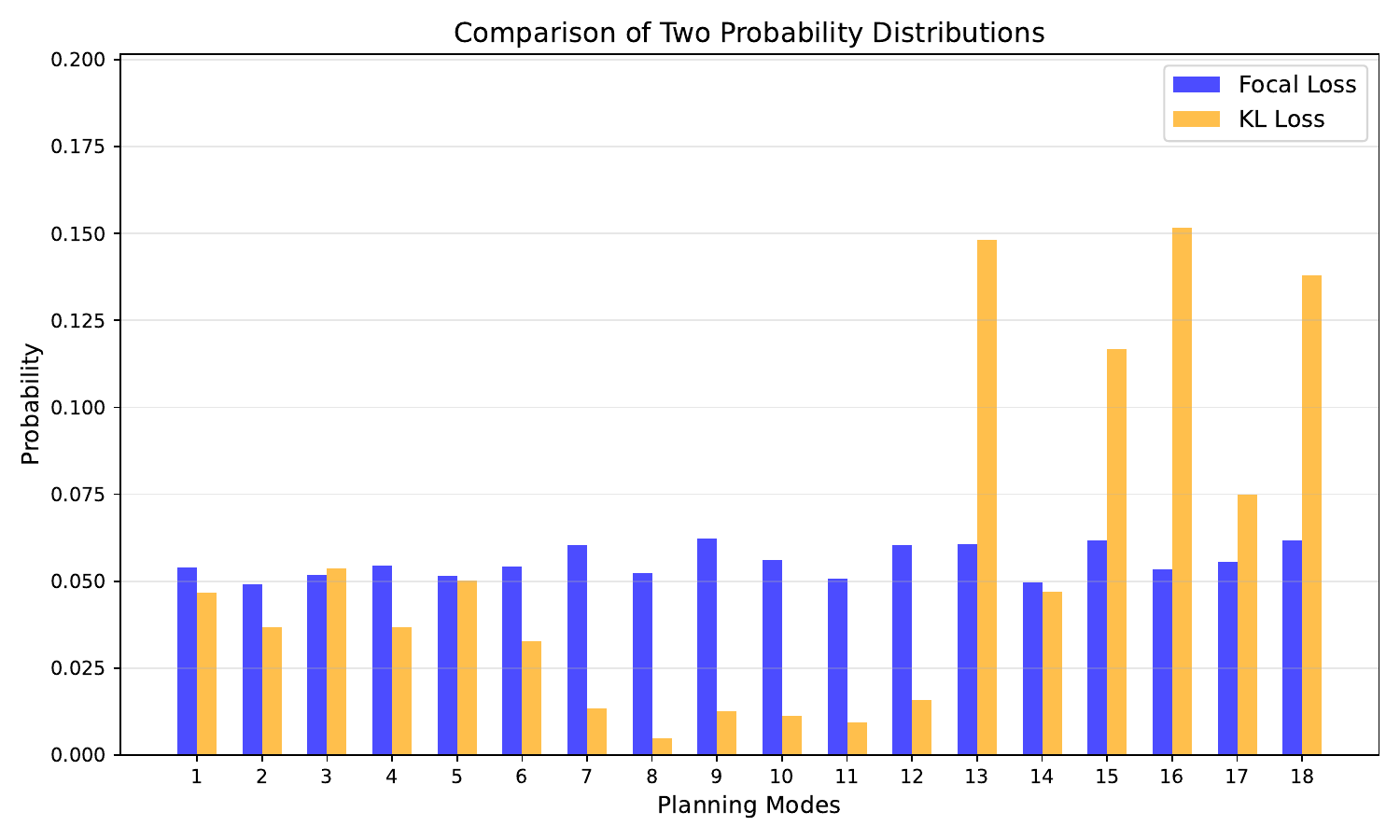}
    \caption{Visualization of probability for target in same scene.}
    \label{figs:fig1}
\end{figure}

\begin{table}[]
\centering
\tiny
    \renewcommand{\arraystretch}{1.1}
    \resizebox{1.0\linewidth}{!}{
    \begin{tabular}{ccc|cc}
    \noalign{\hrule height 0.5pt}
    Distillation           & Generative           & RL  & EP $\uparrow$ & PDMS $\uparrow$ \\ \hline
    \ding{55}  & \ding{55}  & \ding{55} & 78.6  & 83.5    \\
    \ding{51} & \ding{55} & \ding{55} & 80.0 & 85.8    \\
    \ding{51} & \ding{51} & \ding{55} & 80.3  & \bfseries 86.0   \\ 
    \ding{51} & \ding{51} & \ding{51} &  \textbf{80.5}  &  85.5   \\ 
    \noalign{\hrule height 0.5pt}
    \end{tabular}
    }
    \caption{\textbf{Ablation study on the NAVSIM val dataset.}}
    \label{tabs:tab4}
\end{table}

\begin{table}[]
    \renewcommand{\arraystretch}{1.1}
    \resizebox{1.0\linewidth}{!}{
    \begin{tabular}{lll|lll}
    \hline
    \noalign{\hrule height 0.5pt}
    Agent & Temporal & Map & Avg Collision(\%) & Avg L2(m) & Best Epochs \\ \hline
    \ding{51}     &  \ding{55}       & \ding{51}    & 0.111             & 1.3915    & 60          \\
       \ding{55}   & \ding{51}         & \ding{51}    & 0.107             & 0.6632    & 60          \\
    \ding{51}      & \ding{51}       & \ding{51}    & \bfseries 0.066  \textcolor{Green}{\lowercase{-40\%}}           & 
    \bfseries 0.5716 \textcolor{Green}{\lowercase{-60\%}}    & 30          \\ \hline
    \noalign{\hrule height 0.5pt}
    \end{tabular}
    }
    \caption{Ablation study of the planning decoder in the teacher model on nuScenes\cite{nuscenes} val dataset.}
    \label{tabs:tab1}
\end{table}

\begin{table}[]
    \renewcommand{\arraystretch}{1.1}
    \resizebox{1.0\linewidth}{!}{
    \begin{tabular}{ll|lll}
    \hline
    \noalign{\hrule height 0.5pt}
     KL Loss & Focal Loss & Avg Collision(\%) & Avg L2(m) & Best Epochs \\ \hline
    \ding{51} & \ding{55}    & 0.072             & \textbf{0.5497}    & 30          \\
    \ding{55} & \ding{51}    & \textbf{0.061}            & 0.5674    & 20          \\ \hline
    \noalign{\hrule height 0.5pt}
    \end{tabular}
    }
    \caption{Ablation study of the classification loss in the teacher model on nuScenes\cite{nuscenes} val dataset.}
    \label{tabs:tab2}
\end{table}

\begin{table}[]
    \renewcommand{\arraystretch}{1.2}
    \resizebox{1.0\linewidth}{!}{
    \begin{tabular}{llll|ll}
    \hline
    \noalign{\hrule height 0.5pt}
    $L_{reg}^{KD}$             & $L_{cls}^{KD}$             & $L_{en}^{KD}$              & $L_{de}^{KD}$                                   & Avg Collision(\%) & Avg L2(m) \\ \hline
    \ding{51} & \ding{55} & \ding{55} & \ding{55} & 0.183             & 0.5901    \\
    \ding{51} & \ding{51} & \ding{55} & \ding{55} & 0.108             & 0.5856    \\
    \ding{51} & \ding{55} & \ding{51} & \ding{55} & 0.096             & 0.5918    \\
    \ding{51} & \ding{55} & \ding{55} & \ding{51} & \textbf{0.091}             & \textbf{0.5842}    \\ \hline
    \noalign{\hrule height 0.5pt}
    \end{tabular}
    }
    \caption{Ablation Study of the regression distillation on the student model on nuScenes val dataset.}
    \vspace{-0.1in}
    \label{tabs:tab3}
\end{table}

\noindent\textbf{Selection of Planning Decoder.}
To validate the decoders design, we assess their performance in \cref{tabs:tab1}.
As the evaluation mainly focuses on collision and planning, we analyze the Map Decoder and further compare the roles and interactions of the Temporal and Agent Decoders.
Introducing the Agent Decoder enables basic interaction and planning, though with some performance fluctuations.
Replacing it with the Temporal Decoder for sequential agents significantly enhances planning, especially by reducing L2 error.
Using both decoders together enhances planning by balancing agent interactions and temporal information.
This approach reduces collision rates and L2 error while improving training efficiency, requiring half the epochs.

\noindent\textbf{Effect of Classification Target.}
In SparseDrive \cite{sparsedrive}, focal loss is employed for supervised classification between dissimilar modalities.
However, it relies solely on unique labels for supervision, lacking the capability for diverse imitation learning.
As introduced in \cref{eqs:8}, we use KL divergence for supervision, so here we perform a simple performance analysis of the two categorical loss.
As shown in Table \cref{tabs:tab2}, the two models exhibit comparable performance.
However, the model with the KL Loss constraint demonstrates superior performance in terms of L2 metrics, while the model with the Focal Loss constraint achieves a lower collision rate.
Distributions supervised by KL divergence exhibit more diverse probabilities in the scene, influenced by the number of TopKs, while label-supervised distributions have relatively uniform probabilities as show in \cref{figs:fig1}.
Therefore, we prefer the KL divergence-constrained classification layer for diversity.

\noindent\textbf{Ablation Study of Regression Distillation.}
In the main text, we analyze classification distillation as less effective for practical motion planning due to its focus on category differences.
Therefore, in \cref{tabs:tab3}, we additionally provide the effect of regression distillation, which can be found to have a negative effect on collisions when used alone, leading to regression head learning disorientation.
While combining it with other distillation methods provides some mitigation, the overall performance remains inferior to multi-mode instance learning, which avoids the representation space gap between across heads.

\begin{figure}
    \centering
    \includegraphics[width=0.45\textwidth]{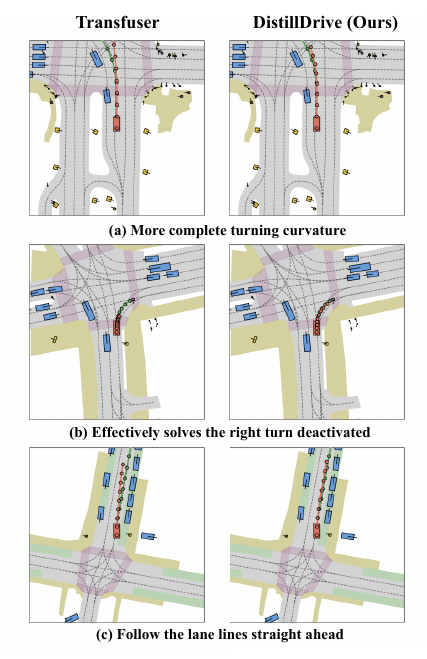}
    \caption{Qualitative visualization of the planning performance on the NAVSIM \cite{NAVSIM} navtest split,follow the logic of the paper.}
    \label{figs:fig3}
\end{figure}

\begin{figure}
    \centering
    \includegraphics[width=0.45\textwidth]{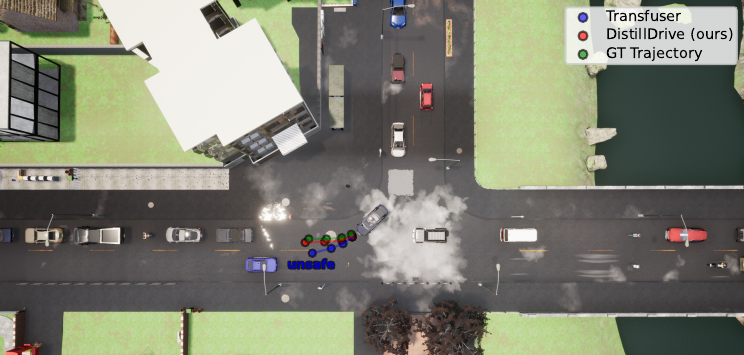}
   \caption{\textbf{Qualitative visualisation on the CARLA dataset.} DistillDrive aligns better with driving demonstrations in turns, making it safer than TransFuser with lower collision rate.}
   \label{figs:fig4}
   \vspace{-0.1in}
\end{figure}

\noindent\textbf{Qualitative Visualization on NAVSIM Dataset.}
In \cref{figs:fig3}, we present additional scenarios from the NAVSIM dataset to illustrate the results.
In case (a), the Transfuser \cite{transfuser} struggles to complete a left turn, whereas DistillDrive successfully assists the model in navigating the turn according to the prescribed curvature.
Similarly, DistillDrive effectively addresses the right-turn deactivated issue of Transfuser, as shown in \cref{figs:fig3} (b).
In the turn-to-straight scenario (c), Transfuser fails to efficiently adjust the yaw, whereas our proposed DistillDrive successfully outputs a planned path that aligns with the lane line as expected.

\noindent\textbf{Qualitative Visualization on CARLA Dataset.}
To better evaluate the model’s performance in closed-loop settings, we provide qualitative comparisons in \cref{figs:fig4}, which show that DistillDrive enhances safety in turns and interactive scenarios through multi-mode distillation and reinforcement learning optimization.

\begin{figure}
    \centering
    \includegraphics[width=0.45\textwidth]{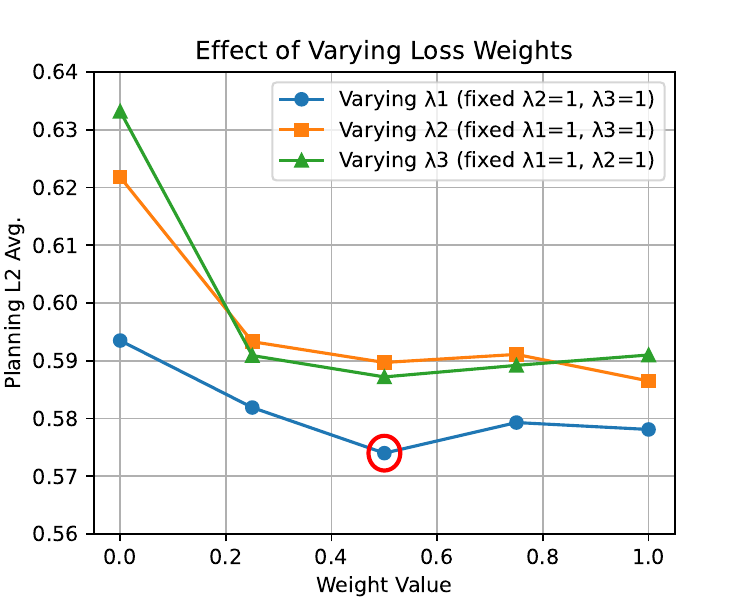}
    \caption{Visualization of loss weight impact on performance.}
    \vspace{-0.1in}
    \label{figs:fig2}
\end{figure}

\noindent\textbf{Ablation Study of Hyperparameters}
We have experimented with the hyperparameter \(\zeta\) of the distributional generative model, and its impact on the model is not significant.
As for the decaying sparse \(\gamma\)  of the reward function in reinforcement learning, we did not do additional experiments and set it to 0.95 by default.
The impact of the loss weight \(\lambda_1, \lambda_2, \lambda_3\) on the model performance is analyzed in \cref{figs:fig2}.
The model achieves optimal performance when \(\lambda_2\) and \(\lambda_3\) are to 1, and \(\lambda_1\) is set to 0.5. Conversely, performance deteriorates significantly when both reinforcement learning and generative models fail, which occurs when \(\lambda_2\) and \(\lambda_3\) are set to 0.
At the same time, the choice of these two parameters (\(\lambda_2\) and \(\lambda_3\)) has minimal impact on the model performance.
 Instead, the multi-mode instance imitation is primarily controlled by the distillation loss weight \(\lambda_1\).

\begin{figure*}[t] 
    \centering
    \includegraphics[width=0.75\textwidth]{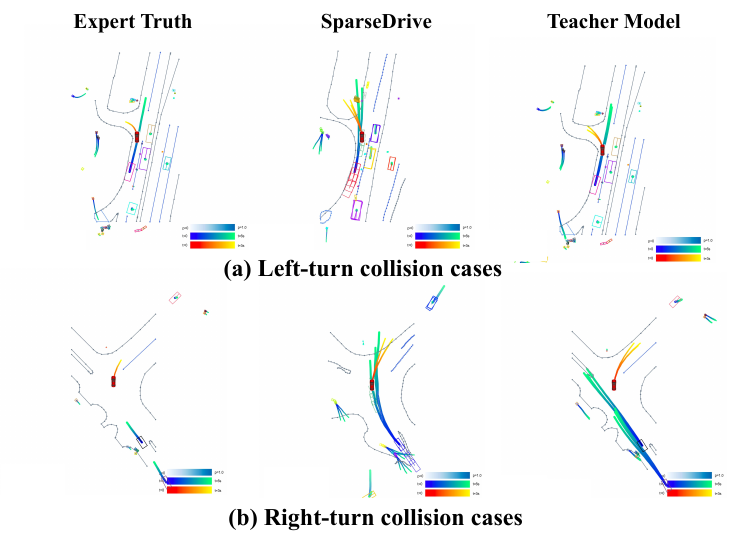}  
    \caption{Qualitative visualization comparing the performance of the teacher model and the end-to-end model SparseDrive, verifying their differences in performance before knowledge distillation.}
    \label{figs:fig5}
\end{figure*}

\begin{figure*}[t] 
    \centering
    \includegraphics[width=0.75\textwidth]{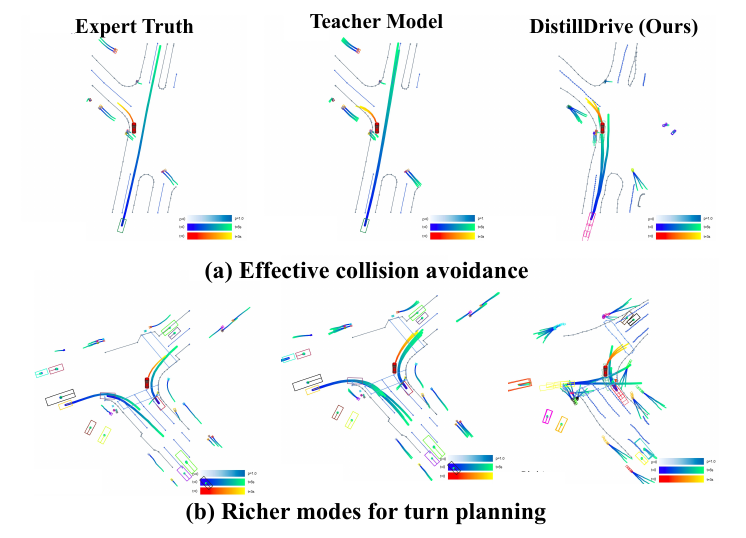}  
    \caption{Qualitative visualization of performance for the teacher model and our proposed end-to-end model DistillDrive.}
    \label{figs:fig6}
\end{figure*}

\noindent\textbf{Effect Analysis of Knowledge Distillation.}
To effectively verify the reasonableness of knowledge distillation, we first visualize our teacher model and SparseDrive in \cref{figs:fig5}.
We observe that SparseDrive frequently encounters issues when turning, often leading to collisions with the road edge.
In contrast, our designed teacher model effectively avoids such situations, further demonstrating its superior ability to perceive lane line information.
Moreover, the designed diverse end-to-end imitation learning model not only captures the distributional representation of multi-modal motion features but also surpasses the teacher model in collision performance in certain cases, as shown in \cref{figs:fig6} (a, b).

\begin{figure*}[t] 
    \centering
    \includegraphics[width=1.0\textwidth]{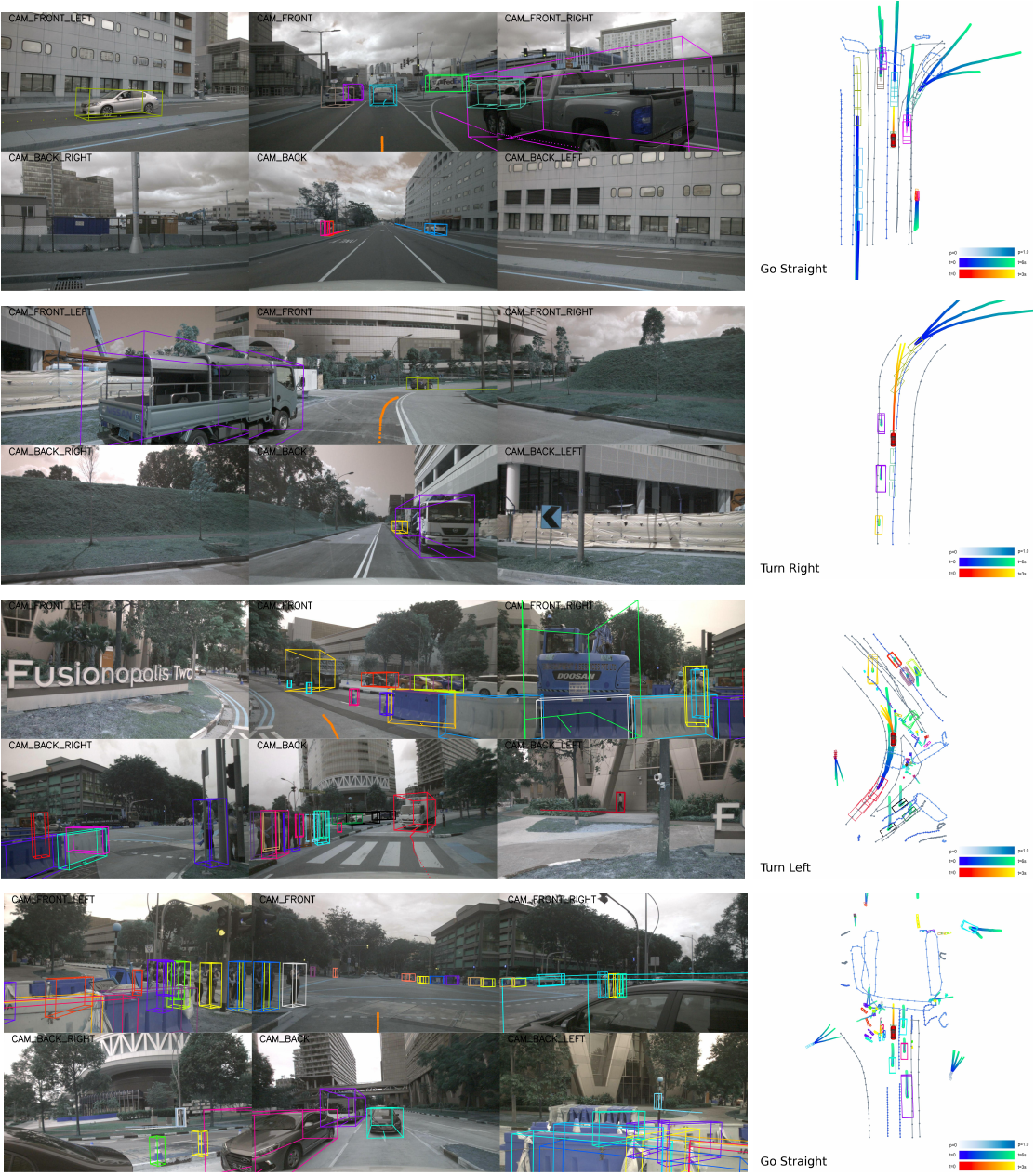}  
    \caption{Qualitative visualization of DistillDrive's overall performance on the nuScenes val dataset.}
    \label{figs:fig7}
\end{figure*}

\noindent\textbf{Qualitative Visualization on nuScenes Dataset.}
In \cref{figs:fig7}, we present a comprehensive visualization of DistillDrive's overall performance on the nuScenes dataset, covering both perception and planning. The results demonstrate the effectiveness of our framework in capturing scene semantics, maintaining temporal consistency, and producing accurate and feasible driving trajectories across diverse urban environments.

\end{document}